\documentclass[letterpaper, 10 pt, conference]{ieeeconf}  

\IEEEoverridecommandlockouts                              

\usepackage{graphicx} 
\usepackage{amsmath} 
\usepackage{amssymb}  
\usepackage{cite}

\usepackage{makecell}  

\title{\LARGE \bf
DA-MMP: Learning Coordinated and Accurate Throwing with Dynamics-Aware Motion Manifold Primitives
}

\author{
Chi Chu$^{1}$, Huazhe Xu$^{2,1}$%
\thanks{%
\protect\raggedright
$^{1}$ Shanghai Qi Zhi Institute; $^{2}$ Institute for Interdisciplinary Information Sciences, Tsinghua University.\protect\\
\protect\hspace*{1em}Contact: cc2018011855@gmail.com, huazhe\_xu@mail.tsinghua.edu.cn}
}

\begin{document}

\maketitle
\thispagestyle{empty}
\pagestyle{empty}

\begin{abstract}
Dynamic manipulation is a key capability for advancing robot performance, enabling skills such as tossing. 
While recent learning-based approaches have pushed the field forward, most methods still rely on manually designed action parameterizations, limiting their ability to produce the highly coordinated motions required in complex tasks. 
Motion planning can generate feasible trajectories, but the dynamics gap—stemming from control inaccuracies, contact uncertainties, and aerodynamic effects—often causes large deviations between planned and executed trajectories. 
In this work, we propose Dynamics-Aware Motion Manifold Primitives (DA-MMP), a motion generation framework for goal-conditioned dynamic manipulation, and instantiate it on a challenging real-world ring-tossing task.
Our approach extends motion manifold primitives to variable-length trajectories through a compact parameterization and learns a high-quality manifold from a large-scale dataset of planned motions. 
Building on this manifold, a conditional flow matching model is trained in the latent space with a small set of real-world trials, enabling the generation of throwing trajectories that account for execution dynamics.
Experiments show that our method can generate coordinated and smooth motion trajectories for the ring-tossing task.
In real-world evaluations, it achieves high success rates and even surpasses the performance of trained human experts.
Moreover, it generalizes to novel targets beyond the training range, indicating that it successfully learns the underlying trajectory–dynamics mapping.
\end{abstract}

\section{INTRODUCTION}

Dynamic manipulation~\cite{mason1993dynamicmanipulation, lynch1999dynamic}---the ability of robots to purposefully interact with objects through high-speed motions---has recently attracted growing interest~\cite{ma2025badminton, su2025hitterhumanoidtabletennis, dambrosio2025achievinghumanlevelcompetitive, zhang2024catchit}. 
While many manipulation tasks can be accomplished under assumptions of negligible object dynamics, dynamic manipulation requires leveraging object motion and momentum to achieve the goal. 
This setting poses two key challenges: coping with intricate, hard-to-model dynamics and generating trajectories that effectively exploit velocity and momentum to achieve complex tasks.

Recent dynamic manipulation systems~\cite{zeng2019tossingbot, chi2022irp, aslam2025dartbot, wang2025ipa} primarily use learning to model or compensate for complex task dynamics, while the motion generation process is typically based on hand-crafted models or manually designed action parameterizations. 
This limits their applicability to more complex tasks that require complicated and highly coordinated motion sequences. 
In dynamic manipulation research, movement primitives (MPs) have been widely used to learn and generate complex motions from demonstrations or planned trajectories~\cite{ijspeert2013dmp, kober2009icra, kober2010icra, paraschos2013promp, zhou2019vmp, pervez2017convdmp, bahl2020neuraldmp, li2023prodmp, saveriano2023servey}. 
By parameterizing motions with a small set of basis-function weights, MPs enable compact representation and efficient modulation of key attributes such as speed, amplitude, and timing, making them particularly suitable for high-speed, dynamic tasks. 
Building on this idea, motion manifold primitives (MMPs)~\cite{noseworthy2022tcvae, lee2023mmp, lee2024mmp, lee2024mmfp, lee2025diffmmp} replace hand-crafted parameter spaces with low-dimensional manifolds learned directly from large collections of feasible trajectories. 
While recent works have extended MMPs to tasks such as tossing~\cite{lee2025diffmmp}, they generally assume fixed-length trajectories, are trained on relatively modest-scale datasets, and have not been fully explored in challenging real-world dynamic manipulation scenarios.

\begin{figure}[t]
    \centering
    \includegraphics[width=\linewidth]{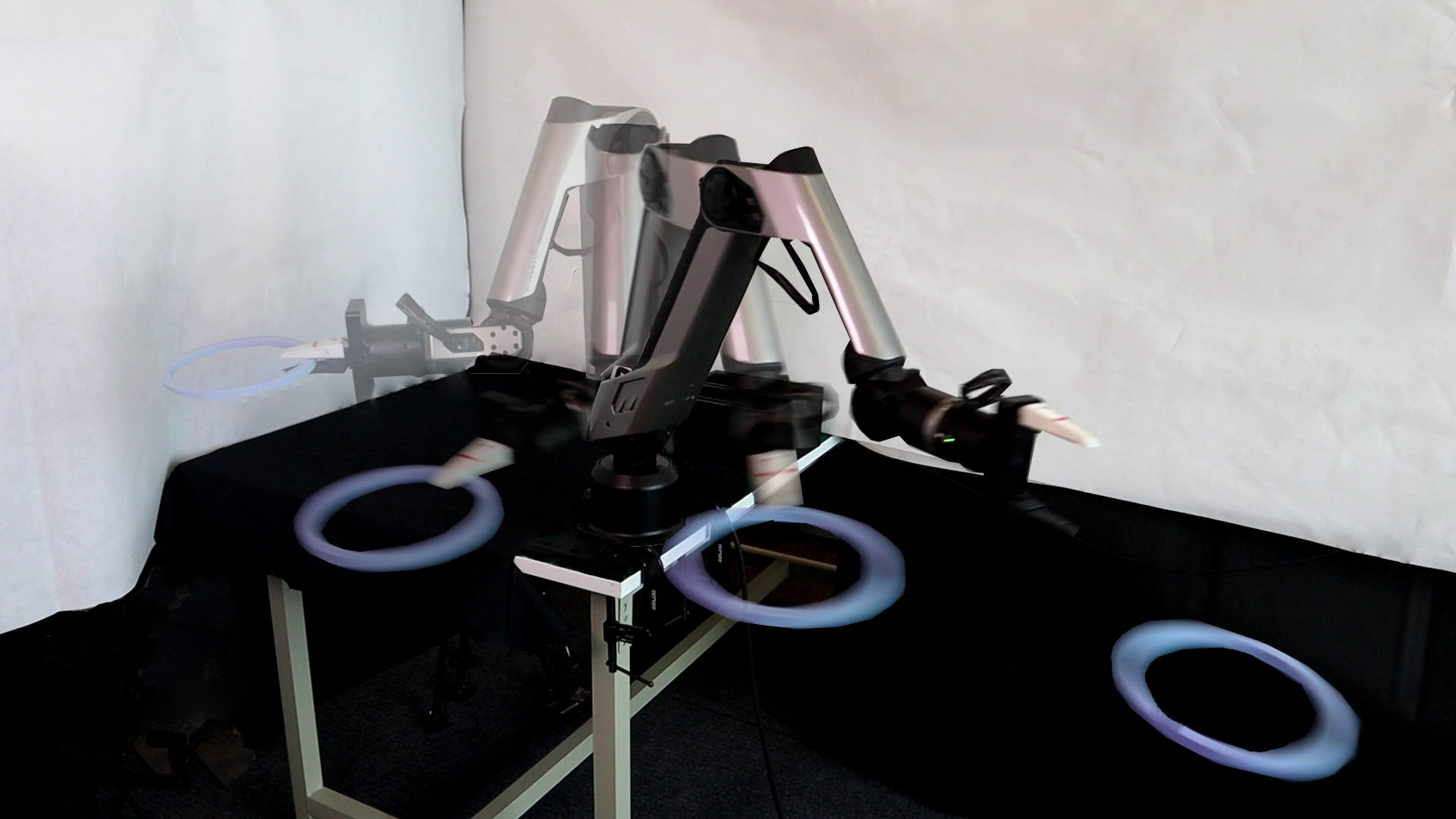} \\[3pt]
    \begin{minipage}{0.325\linewidth}
        \centering
        \includegraphics[width=\linewidth]{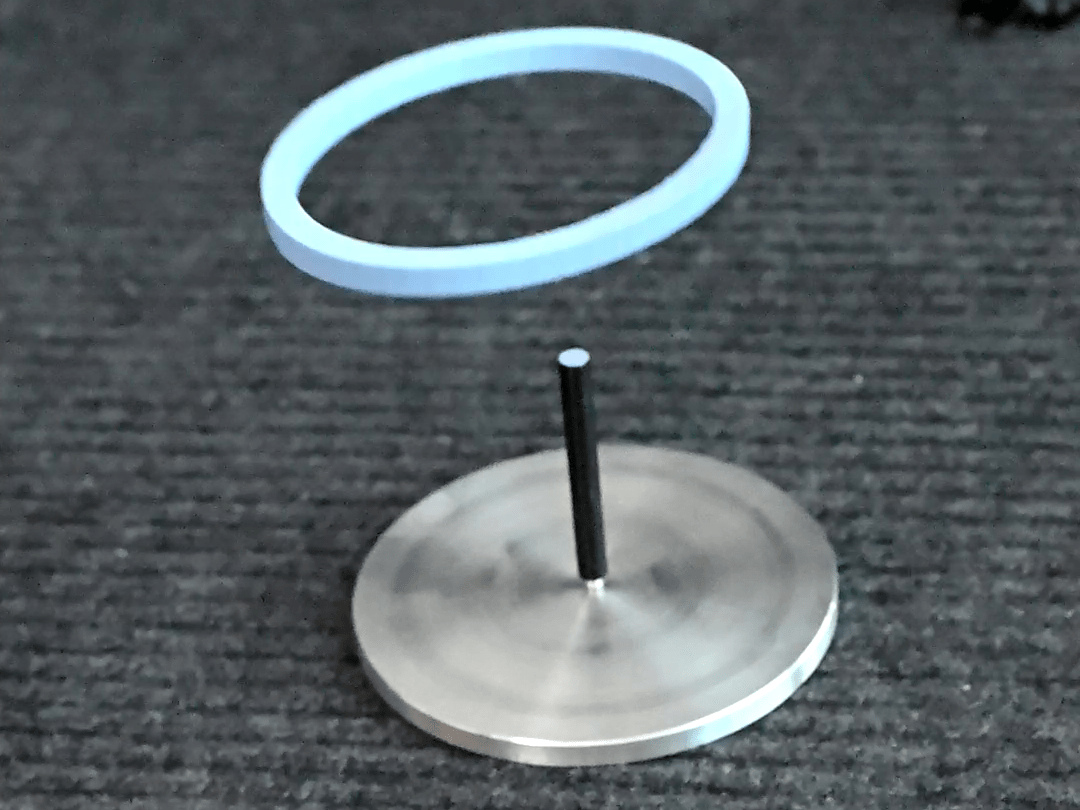}
    \end{minipage}
    \hfill
    \begin{minipage}{0.325\linewidth}
        \centering
        \includegraphics[width=\linewidth]{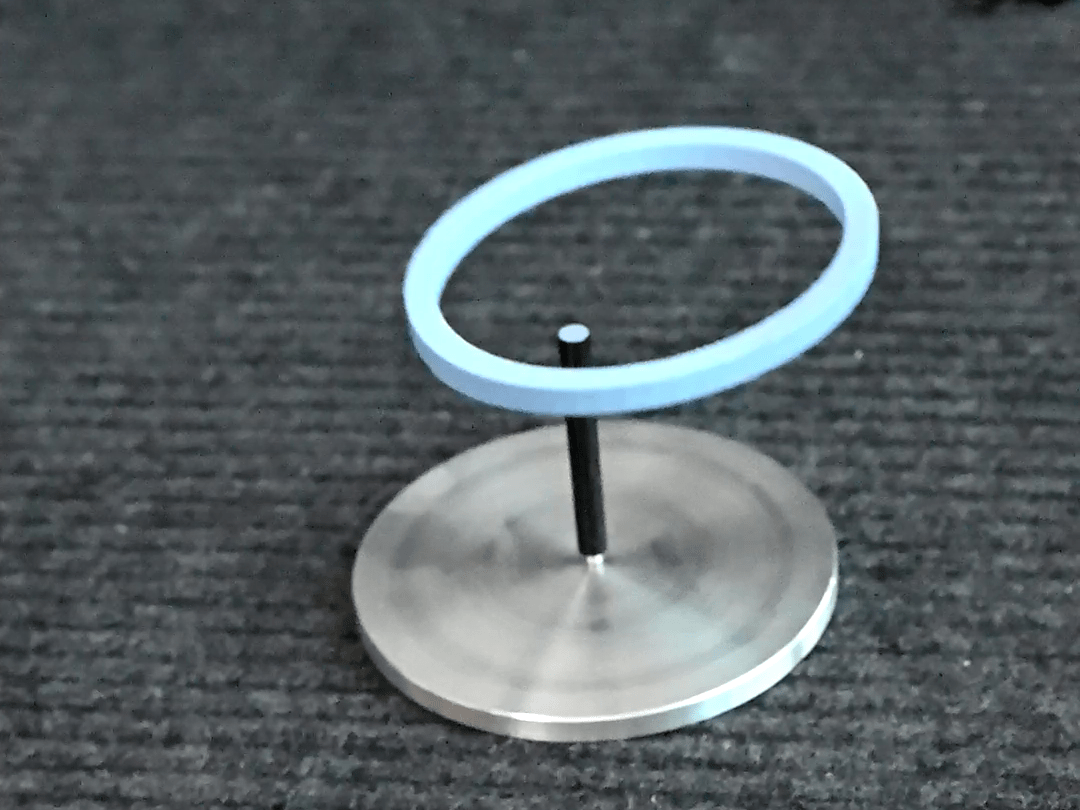}
    \end{minipage}
    \hfill
    \begin{minipage}{0.325\linewidth}
        \centering
        \includegraphics[width=\linewidth]{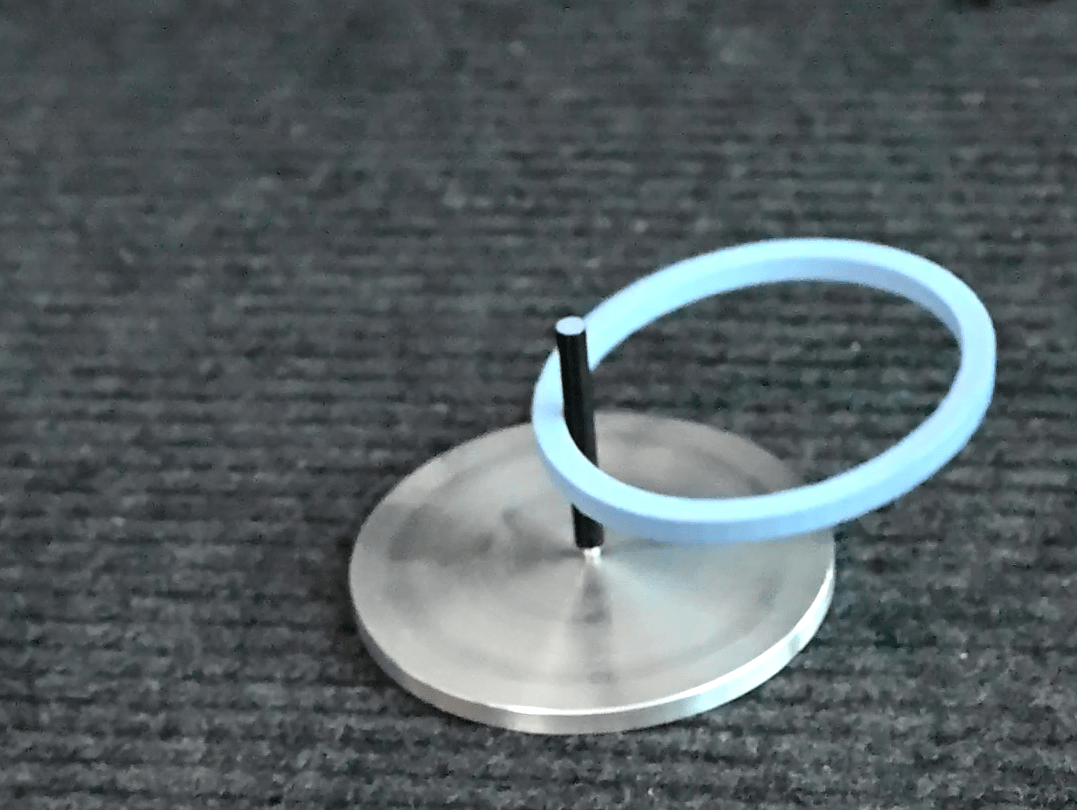}
    \end{minipage}
    \caption{Real-world ring tossing with trajectories synthesized by DA-MMP. The first line shows the overall sequence, where superimposed snapshots (light to dark) indicate temporal order of the coordinated throw. The second line illustrates three representative frames of the hitting process.}
    \label{fig:ring_toss}
\end{figure}

In this work, we bridge these two lines of research by introducing Dynamics-Aware Motion Manifold Primitives (DA-MMP), a motion generation framework for complex dynamic manipulation that integrates motion manifold primitives with large-scale planning data and trajectory-level dynamics learning.
We focus on a challenging real-world ring-tossing task, as illustrated in Fig.~\ref{fig:ring_toss}, that requires generating highly coordinated motions to throw a ring around a distant target peg.
This setting is particularly difficult due to multiple sources of dynamics gaps, including control inaccuracies of the arm (control gaps), slippage and uncertainty in release timing (contact gaps), and aerodynamic effects during flight (aerodynamic gaps).
Our framework extends MMPs to variable-length trajectories via a compact parameterization that includes execution duration, and leverages a large-scale dataset of planned trajectories generated by sampling-based motion planning to learn a high-quality motion manifold.
A conditional generative model is then trained in the learned latent space to generate throwing trajectories. 
By training on the observed landing outcomes of executed trajectories, this model implicitly captures the relationship between trajectories and their execution dynamics.
Experiments demonstrate that our approach achieves high real-world success rates, even rivaling trained human experts, and can generalize to unseen targets, which indicates that it generates valid trajectories and effectively captures task dynamics.

Our work makes the following contributions at the intersection of dynamic manipulation and motion manifold learning:%
\begin{itemize}
    \item \textbf{Challenging dynamic manipulation task.} We design and address a real-world ring-tossing task that requires generating complex, highly coordinated motions, providing a demanding benchmark for evaluating expressive motion generation.
    \item \textbf{Variable-length motion manifold primitives with dynamics learning.} We extend the MMP framework to represent variable-length trajectories through a compact parameterization that includes execution duration, and introduce a conditional generative model in the latent space that captures the relationship between trajectories and their execution dynamics. 
    \item \textbf{Large-scale dataset and efficient real-world learning.} Leveraging sampling-based motion planning, we collect a large-scale dataset of feasible trajectories to learn a high-quality motion manifold, and further enable data-efficient learning from limited real-world trials.
\end{itemize}

\section{Related Work}

Our approach builds on and connects two main lines of research. 
The first line of research is motion manifold primitives, which learn low-dimensional latent representations of feasible trajectories. 
The second line is learning-based goal-conditioned dynamic manipulation, which applies machine learning to address the challenges of high-speed, target-directed manipulation. 
We review both directions in the following subsections.

\subsection{Motion Manifold Primitives}

Motion Manifold Primitives~\cite{lee2023mmp} build upon the idea of movement primitives~\cite{ijspeert2013dmp, paraschos2013promp, zhou2019vmp}, 
which offer structured representations for generating robot motions. 
Instead of relying on hand-crafted parameter spaces, MMPs learn low-dimensional manifolds directly from large collections of feasible trajectories. 
A closely related idea was explored by Noseworthy et al.~\cite{noseworthy2022tcvae}, who used task-conditioned variational autoencoders to learn latent spaces of movement primitives. 
Subsequent work has expanded MMPs in several directions, including conditional generative models for task-specific motion synthesis~\cite{lee2024mmfp}, 
extensions with parametric curves for temporal and via-point modulation~\cite{lee2024mmp}, 
and a differentiable formulation that enables further optimization of the learned manifold with downstream objectives~\cite{lee2025diffmmp}.

While these efforts have demonstrated MMPs across diverse domains, they typically assume fixed-length trajectories, are trained on relatively modest-scale datasets, and have not been fully explored in challenging real-world dynamic manipulation. In contrast, our work extends MMPs to variable-length trajectories via a compact parameterization, learns high-quality manifolds from large-scale planned motion data without requiring downstream post-optimization, and incorporates a conditional generative model that implicitly captures the relationship between trajectories and their execution dynamics.

\subsection{Learning-based goal-conditioned dynamic manipulation}

Recent works have applied deep learning to high-speed, goal-conditioned manipulation tasks with intricate dynamics. 
Among the extensive literature, we highlight several representative efforts. 
TossingBot~\cite{zeng2019tossingbot} learns residual release velocities conditioned on grasp and object's visual features to compensate for various dynamics gaps. 
IRP~\cite{chi2022irp} addresses dynamic manipulation of deformable objects by learning delta dynamics in an iterative residual policy framework, enabling online adaptation to new dynamics and hardware. 
DartBot~\cite{aslam2025dartbot} leverages tactile sensing and reinforcement learning to exploit contact interactions, learning slippage and release timing in real-world overhead dart throwing. 
Whole-body throwing with legged manipulators~\cite{ma2025learningaccuratewholebodythrowing} combines a residual control policy to address control gaps with tube acceleration~\cite{liu2024tubeacceleration} to handle release-time uncertainties. 
The IPA policy~\cite{wang2025ipa} further tackles dynamic manipulation with soft-body tools by implicitly identifying the underlying physics of soft–rigid interactions through system identification.
Unlike these approaches, which largely rely on hand-crafted motion generation rules with constrained expressiveness, our method synthesizes high-dimensional trajectories through planning and manifold learning.
Moreover, target-level residual learning, as in TossingBot, is unsuitable in our setting since distinct trajectories for the same target can yield different landings; instead, we learn dynamics directly at the trajectory level to capture such differences.

Like our work in spirit, earlier studies also applied movement primitives to dynamic manipulation to generate smooth and consistent trajectories. 
Representative examples include learning motor primitives for hitting and batting~\cite{kober2010icra}, swinging and padding~\cite{kober2009icra}, and applying mixtures of motor primitives to table tennis~\cite{muelling2010mixtureofmotorprimitives, mulling2013ijrr}.
These approaches relied on adapting from a limited number of demonstrations, whereas our method learns motion manifold primitives from large-scale planned trajectories and uses limited real-world data to capture dynamics.


\begin{figure*}[t]
    \centering
    \includegraphics[width=\linewidth, trim=0 140 0 140, clip]{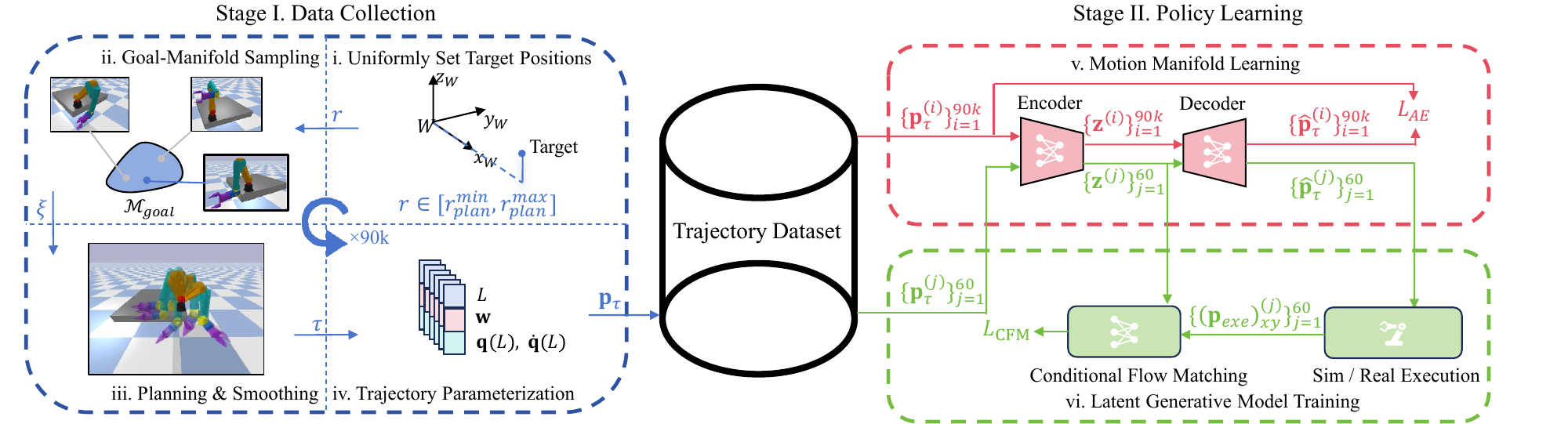}
    \caption{Overview of DA-MMP. 
    Stage~I (blue): data collection from goal-manifold sampling, motion planning, and parameterization, yielding $90$k planned trajectories. 
    Stage~II (red/green): policy learning with an autoencoder to learn motion manifolds and a conditional flow-matching model for trajectory-dynamics mapping.}
    \label{fig:method_overview}
\end{figure*}

\section{Method}

In this section, we present DA-MMP, a framework that leverages motion manifold learning for dynamics-aware trajectory generation. The framework consists of two stages, as shown in Fig.~\ref{fig:method_overview}: (1) data collection and parameterization, and (2) policy learning using a motion manifold representation and a conditional flow-matching model. We begin by introducing the problem formulation, followed by a detailed description of each stage.

\subsection{Problem Formulation}
We study goal-conditioned dynamic manipulation tasks that require complex motions, exemplified by a ring-tossing task. 
In this task, the objective is to throw a ring of radius $R_{\mathrm{ring}}$ such that it lands around a vertical cylinder of radius $R_{\mathrm{cyl}}$ and height $z_{\mathrm{cyl}}$ on the ground. 
To simplify the problem, we assume a fixed hand-crafted grasp of the ring. 
A tossing trajectory is defined as 
\begin{equation}
    \tau = \{ \mathbf{q}(t), \dot{\mathbf{q}}(t) \}_{t=0}^L ,
\end{equation}
where $\mathbf{q}(t)$ and $\dot{\mathbf{q}}(t)$ denote the joint positions and velocities at time $t$, and $L$ is the execution horizon, which can vary across different trajectories. 
At time $t = L$, the gripper starts opening. 
During a subsequent settling horizon $h_{\mathrm{settle}}$, the end-effector maintains its linear and angular velocity, 
allowing sufficient time for the ring to be released from the gripper. 
Inspired by how humans perform ring tossing, we design the robot to start and end at the same joint configuration with zero initial velocity:
\begin{equation}
    \mathbf{q}(0) = \mathbf{q}(L), \quad \dot{\mathbf{q}}(0) = \mathbf{0}.
\end{equation}

Given the target position, we exploit the rotational symmetry of the manipulator’s first joint and represent the target in polar coordinates. 
Without loss of generality, we fix the polar angle $\alpha=0$, so that the problem reduces to generating a trajectory $\tau(r)$ for a given radial distance $r$. In practice, however, the target placement and the ring landing position are not perfectly aligned with $\alpha=0$, so we represent both using their $(x,y)$ coordinates, where $y$ is nearly zero.
We consider a trial as successful if, at height $z=z_{\mathrm{cyl}}$, 
\begin{equation}
  \big\| (x_{\mathrm{ring}}(z_{\mathrm{cyl}}),\, y_{\mathrm{ring}}(z_{\mathrm{cyl}})) - (x_T, y_T) \big\|_2
  \;\le\; R_{\mathrm{ring}} - R_{\mathrm{cyl}} ,
\end{equation}
where $(x_{\mathrm{ring}}(z_{\mathrm{cyl}}),\, y_{\mathrm{ring}}(z_{\mathrm{cyl}}))$ are the horizontal coordinates of the ring center at $z=z_{\mathrm{cyl}}$, and $(x_T, y_T)$ denotes the horizontal coordinates of the target top center.
This serves as an approximation, since the ring may not remain perfectly horizontal during flight.

\subsection{Data Collection}
\subsubsection{Goal-Manifold Sampling}
We adopt the idea of goal-manifold sampling from prior work on robotic throwing~\cite{zhang2012icra, pekarovskiy2013goalmanifold}, 
where projectile equations are used to generate candidate throw states $\xi$, forming a manifold of feasible goals $\mathcal{M}_{\text{goal}}$ used for motion planning. 
Each candidate specifies the desired end-effector (EE) pose and velocity at release.
If the planner fails to generate a feasible trajectory, a new candidate is drawn until success.

Formally, we parameterize each throw state in the ring space, since the EE state is uniquely determined via a constant grasp transform ${}^E T_R$. 
The grasp is designed such that the ring center lies along the EE $z$-axis and the release direction aligns with the EE $x$-axis, which reduces the likelihood of post-release collisions with the gripper.
We denote a ring throw state by $\xi_R$, consisting of its pose and twist in the world frame:
\[
\xi_R = \big({}^W T_R,\; {}^W \mathbf{v}_R,\; {}^W \boldsymbol{\omega}_R\big) \in SE(3)\times\mathbb{R}^6,
\]
which comprises twelve degrees of freedom. 
In the following we describe the constraints that define this feasible manifold of throws.

\begin{figure}[t]
  \centering
  \includegraphics[width=\linewidth, trim=280 150 280 150, clip]{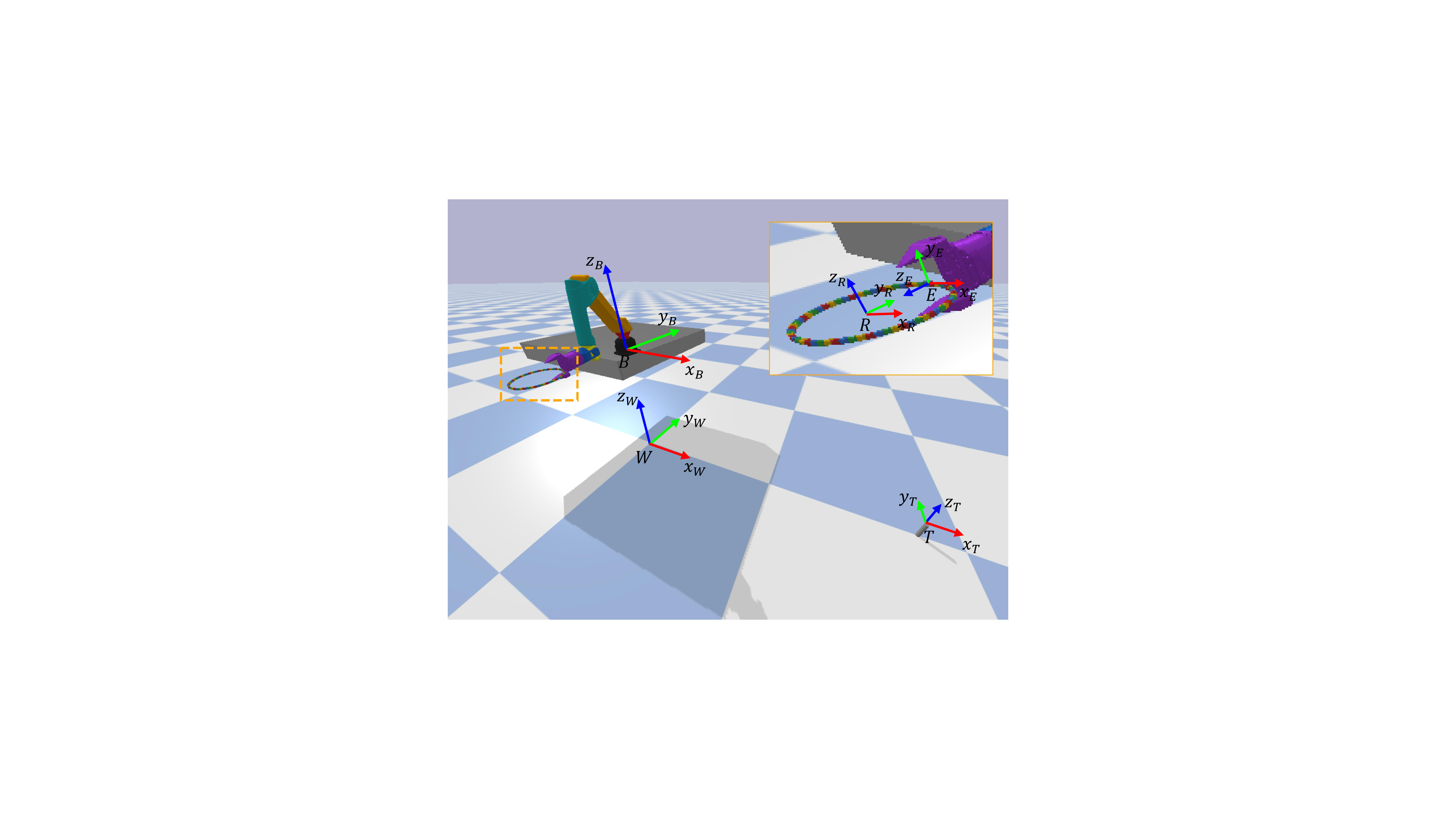}
  \caption{Coordinate frames used for goal-manifold sampling. 
  World $W$ is fixed to the ground; $B$ attaches to the robot base; 
  the end-effector $E$ rigidly grasps the ring frame $R$ via ${}^E T_R$; 
  the target frame $T$ sits at the cylinder top center. 
  Inset: enlarged view of $E$ and $R$ for clarity.}
  \label{fig:frames}
\end{figure}

The twelve degrees of freedom of $\xi_R$ are constrained as follows, with reference to the coordinate frames in Fig.~\ref{fig:frames}. 
Let ${}^W\mathbf{p}_R$ and ${}^W\mathbf{p}_T$ denote the positions of the ring center and the target top center in the world frame $W$. 
The orientation is fixed by geometry, where $(x_R, y_R, z_R)$ denote the unit axes of the ring frame $R$:
\[
x_R \parallel ({}^W\mathbf{p}_T - {}^W\mathbf{p}_R)_{xy}, 
\qquad z_R \parallel z_W, 
\qquad y_R = z_R \times x_R .
\]
The angular velocity is restricted to spin about the ring normal:
\[
{}^W\boldsymbol{\omega}_R = (0,\,0,\,\omega_z)^\top,
\]
where $\omega_z$ is sampled from the range $[\omega_{\min},\,\omega_{\max}]$. 
The ring center position ${}^W \mathbf{p}_R=(x,y,z)$ is sampled within the workspace.
In practice we use a set of surrogate variables to generate these samples, but omit the details here for simplicity.
For linear velocity, we impose
\[
v_z = 0, \qquad {}^W \mathbf{v}_R \parallel x_R ,
\]
where $v_z$ denotes the vertical component of ${}^W \mathbf{v}_R$.
Its magnitude is determined by projectile motion to reach the target height $z=z_{\mathrm{cyl}}$:
\begin{equation}
\label{eq:proj-speed-final}
\|{}^W \mathbf{v}_R\| = d_{xy}\,\sqrt{\tfrac{g}{2\,(z_R-z_{\mathrm{cyl}})}} ,
\end{equation}
where $d_{xy}=\|({}^W\mathbf{p}_T-{}^W\mathbf{p}_R)_{xy}\|$ is the horizontal distance to the target, and $g$ is the gravitational acceleration constant.
Together these constraints define a low-dimensional manifold of throws. 
In practice, candidates are drawn from this manifold and checked for feasibility using inverse kinematics and collision detection; if a candidate fails, sampling is repeated until a feasible one is found.

\subsubsection{Sampling-Based Motion Planning}
Sampling-based motion planning is renowned for its computational efficiency in high-dimensional spaces~\cite{lavalle2006planning}, which enables us to generate a large-scale dataset of feasible trajectories.
Given a sampled throw state, we construct the motion planning problem by setting the goal state to this specified throw state and the initial state to the same joint configuration with zero velocity. 
We employ DIMT-RRT as our kinodynamic planning algorithm~\cite{kunz2014dimtrrt}, 
where each planning attempt is allowed up to $N_{\mathrm{planning}}$ samples. 
Successful plans are further refined with $N_{\mathrm{smoothing}}$ iterations of a trajectory smoothing algorithm based on~\cite{hauser2010smoothing}. 
We also adopt the quick rejection strategy proposed in~\cite{zhang2012icra} to efficiently discard infeasible candidates before planning. 
During planning, we perform discrete collision checking with a time resolution of $\Delta t_{\mathrm{col}}$. 

After a trajectory is found, we execute it twice in simulation and compare the resulting landing positions; 
the trajectory is retained only if the deviation between the two trials is below a threshold, 
thus filtering out unstable throwing trajectories. Finally, the accepted trajectory is interpolated to match the control frequency $f_{\mathrm{ctrl}}$.

\subsection{Policy Learning}
\subsubsection{Variable-Length Parameterization of Motion Trajectories}
A key challenge in trajectory parameterization is that planned trajectories have variable durations and cannot be directly resampled or aligned,
since naive interpolation effectively accelerates or decelerates the motion and alters the release velocity that defines the throw state.
To handle variable durations without distorting the motion,
we include the trajectory length $L$ as an explicit component of the trajectory parameterization.
Another challenge is ensuring that the generated trajectories are sufficiently smooth for high-speed execution,
which we address by drawing inspiration from movement primitives and representing trajectories with radial basis functions.
Formally, each trajectory dimension is approximated as a weighted combination of normalized Gaussian radial basis functions,
\begin{equation}
\label{eq:rbf-mp}
\begin{aligned}
q(s;\mathbf{w}) &= \mathbf{w}^\top \boldsymbol{\phi}(s), \\
\phi_i(s) &= \frac{\kappa_i(s)}{\sum_{j=1}^{K} \kappa_j(s)}, \qquad i=1,\dots,K, \\
\kappa_i(s) &= \exp\!\left(-\frac{(s - c_i)^2}{h}\right),
\end{aligned}
\end{equation}
where $s \in [0,1]$ is the normalized phase, $c_i$ are the basis centers, $h$ is the bandwidth, and $\mathbf{w}$ are the basis weights for this trajectory dimension.
Here $\boldsymbol{\phi}(s) = [\phi_1(s), \dots, \phi_K(s)]^\top$ denotes the vector of normalized basis functions.

\begin{figure}[t]
    \centering
    \includegraphics[width=\linewidth]{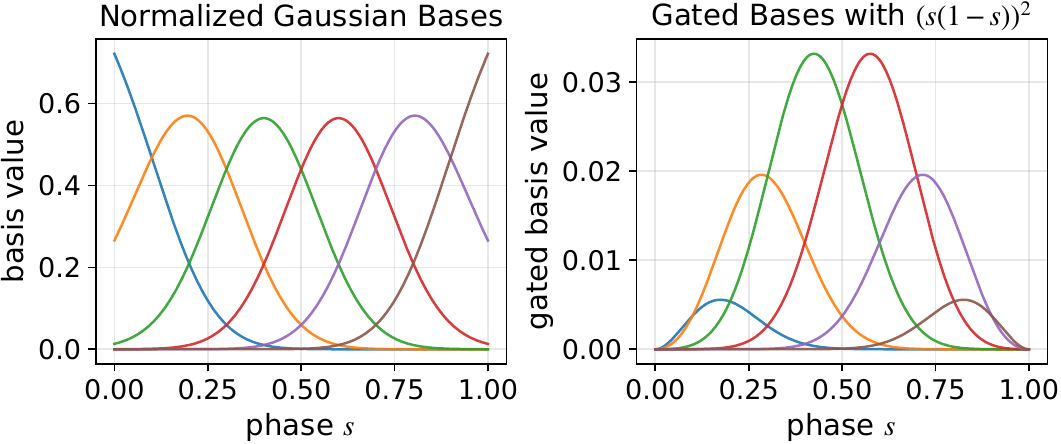}
    \caption{
    Illustration of basis functions used for trajectory parameterization. 
    Left: normalized Gaussian bases over the phase. 
    Right: gated bases obtained by multiplying the normalized bases with the polynomial gate $(s(1-s))^2$. 
    For illustration, we show $K=6$.
    }
    \label{fig:basis_functions}
\end{figure}

We find that directly using radial basis functions often leads to oscillations in the end-point velocity.
To address this, we adopt via-point movement primitives~\cite{zhou2019vmp}, which explicitly constrain the start- and end-point positions and velocities.
Formally, each trajectory is expressed as
\begin{equation}
    q(s;\mathbf{w}) = \psi(s) + \mathbf{w}^\top \boldsymbol{\phi}(s),
\end{equation}
where $\psi(s)$ is a cubic Hermite spline interpolating the boundary conditions 
$\{q(0), q(1), \dot q(0), \dot q(1)\}$,
and $\boldsymbol{\phi}(s)$ denotes the gated basis functions, obtained by multiplying the normalized Gaussian bases with the polynomial gate function $(s(1-s))^2$. For illustration, we show examples of the normalized Gaussian and gated bases in Fig.~\ref{fig:basis_functions}.

Given a reference trajectory defined on $L$ phase points 
$\{(q(s_l), \dot q(s_l))\}_{l=1}^L$ with $s_l = \tfrac{l}{L}$,
the weights $\mathbf{w}$ are obtained by least-squares fitting with equal weighting of position and velocity terms,
\begin{equation}
\label{eq:vmp-ls}
\begin{aligned}
\mathbf{w}^\star = \arg\min_{\mathbf{w}} \;
& \sum_{l=1}^{L} \big( q(s_l) - \psi(s_l) - \boldsymbol{\phi}(s_l)^\top \mathbf{w} \big)^{2} \\
&+ \sum_{l=1}^{L} \big( \tfrac{1}{\alpha}\dot q(s_l) - \psi'(s_l) - \boldsymbol{\phi}'(s_l)^\top \mathbf{w} \big)^{2},
\end{aligned}
\end{equation}
where $\psi(s_l)$ and $\psi'(s_l)=\tfrac{d\psi}{ds}(s_l)$ denote the spline and its phase derivative,
$\boldsymbol{\phi}(s_l)$ and $\boldsymbol{\phi}'(s_l)=\tfrac{d\boldsymbol{\phi}}{ds}(s_l)$ are the gated basis functions and their phase derivatives evaluated at $s_l$,
and $\alpha=\tfrac{ds}{dt}=\tfrac{1}{L}$ is the time-to-phase scaling factor (so $(1/\alpha)\dot q = dq/ds$).
Each trajectory is therefore represented by weights $\mathbf{w}$, the end-point conditions $(q(1), \dot q(1))$, and the trajectory length $L$, 
which we collectively denote as the parameter vector $\mathbf{p}_\tau$.

\begin{figure}[t]\centering
  \includegraphics[width=\linewidth, trim=300 100 300 100, clip]{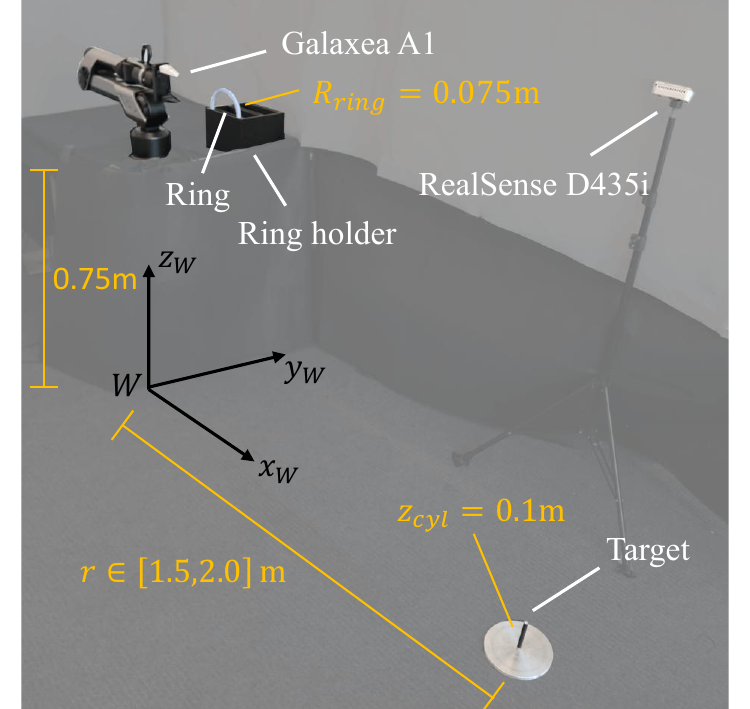}
  \caption{Experimental setup of real-world ring tossing.}
  \label{fig:real_setup}
\end{figure}

\subsubsection{Learning Motion Manifold Primitives}
Given the trajectory parameterization $\mathbf{p}_\tau$, we learn motion manifold primitives by first embedding trajectories into a low-dimensional latent space $\mathbf{z}_\tau$ using an autoencoder, which captures the underlying manifold structure and reduces dimensionality.
The autoencoder is trained to minimize the reconstruction loss
\begin{equation}
    \mathcal{L}_{\mathrm{AE}} = \frac{1}{N}\sum_{i=1}^N \big\| \mathbf{p}_\tau^{(i)} - \hat{\mathbf{p}}_\tau^{(i)} \big\|_2^2 ,
\end{equation}
where $\hat{\mathbf{p}}_\tau^{(i)}$ is the decoded reconstruction of $\mathbf{p}_\tau^{(i)}$. 

On top of this latent representation, we train a conditional flow-matching model~\cite{lipman2023flowmatchinggenerativemodeling, liu2022rectifiedflow, albergo2024stochasticinterpolants, lipman2024flowmatchingguide}, 
where the conditioning signal is given by the actual landing point $(x_{\mathrm{exe}}, y_{\mathrm{exe}})$ of the executed trajectory measured at $z=z_{\mathrm{cyl}}$, implemented via classifier-free guidance~\cite{ho2022classifierfreediffusionguidance}. 
The model is optimized with the standard conditional flow-matching loss
\begin{equation}
\mathcal{L}_{\mathrm{CFM}} 
= \mathbb{E}_{u,\,\mathbf{z}_\tau,\,\mathbf{z}_{\text{noise}}} 
\big[ \| v_\theta(\mathbf{z}(u), u, \mathbf{c}) - v^\star(u) \|_2^2 \big],
\end{equation}
where $u \in [0,1]$ denotes the flow time parameter, 
$\mathbf{z}(u)$ is the linear interpolation between a Gaussian prior sample $\mathbf{z}_{\text{noise}}$ and the ground-truth latent variable of the trajectory $\mathbf{z}_\tau$, 
$v^\star(u)$ denotes the corresponding target velocity field, 
and $\mathbf{c}=(x_{\mathrm{exe}}, y_{\mathrm{exe}})$ is the conditioning signal.

At inference time, given the conditioning signal $\mathbf{c}=(x_T, y_T)$, a noisy latent sample $\mathbf{z}_{\text{noise}}$ is progressively transported to a trajectory latent $\mathbf{z}_\tau$ by integrating the learned vector field $v_\theta(\mathbf{z}(u), u, \mathbf{c})$ over $u \in [0,1]$. The resulting latent $\mathbf{z}_\tau$ is then decoded by the autoencoder to obtain the parameterized trajectory, which is used for the ring-tossing task.

This design implicitly captures the dynamics gap between planned and executed motions. 
Notably, the autoencoder is trained on a large set of planned trajectories to learn a high-quality motion manifold, 
while the flow-matching model is learned in the latent space and thus requires only a modest amount of real-world data to capture trajectory structure and dynamics information.

\begin{table}[t]
\centering
\caption{
Success rate (SR) comparison of different baselines and our method in simulation and real-world experiments.
}
\renewcommand{\arraystretch}{1.1}
\setlength{\tabcolsep}{8pt}
\resizebox{\columnwidth}{!}{%
\begin{tabular}{lcc}
\Xhline{1.2pt}
\textbf{Method} & \textbf{Simulation SR (\%)} & \textbf{Real SR (\%)} \\
\hline
Motion planning (1 attempt)  & 0.0   & 13.3 \\
Motion planning (2 attempts) & 0.0   & 23.3 \\
Residual-style correction    & \textbf{93.3} & 6.7 \\
DA-MMP (Ours)                & 73.3  & \textbf{60.0}$^*$ \\
Human novice                 & --    & 13.3 \\
Human expert                 & --    & 56.7 \\
\Xhline{1.2pt}
\multicolumn{3}{p{\columnwidth}}{\vspace{2pt}\scriptsize $^*$Evaluated on an early implementation version where the trajectory parameterization inadvertently omitted a coefficient (magnitude $\sim 1$) related to the time-to-phase scaling factor in two terms, causing the executed trajectories to deviate from the planned ones. However, since our method trains on actual executed landing positions rather than planned targets, the model is robust to this systematic bias as long as it does not severely distort the motion or violate hardware safety limits. Correcting this omission in simulation improved the success rate from 53.3\% to 73.3\%, while the real-world result retains the early evaluation due to hardware setup deprecation.}
\end{tabular}
}
\label{tab:dynamics_aware}
\vspace{-10pt}
\end{table}

\section{EXPERIMENTS}

\subsection{Setup}
\subsubsection{Simulation and Real-World Setup}
We evaluate our approach in both simulation and real-world experiments using a 6-DoF Galaxea A1 fixed-base robot arm, 
with simulation implemented in PyBullet~\cite{coumans2021pybullet}. 
The arm is mounted on a fixed table of height $0.75$\,m, providing sufficient flight time for the ring. Its joint acceleration limits are set to $[12.5, 12.5, 12.5, 15.0, 15.0, 15.0]$\,m/s$^2$ to balance safety and motion capability. 
The ring has a radius of $R_{\mathrm{ring}} = 0.075$\,m, and the target is a vertical cylinder of radius $R_{\mathrm{cyl}} = 0.005$\,m and height $z_{\mathrm{cyl}} = 0.1$\,m placed on the ground. 
In the real setup (Fig.~\ref{fig:real_setup}), a RealSense D435i depth camera is used to detect the $x,y$ coordinates of the target and the ring position at $z = 0.1$\,m. For the ring, contours are extracted by OpenCV Canny edge detection and fitted with ellipses for robust localization~\cite{opencv_library}.

\subsubsection{Data Collection Details}
We collect a dataset of 90k feasible trajectories for training the autoencoder, 
with target radial distances uniformly sampled within $[r_{\mathrm{plan}}^{\min}, r_{\mathrm{plan}}^{\max}] = [1.0, 2.5]$\,m. 
For goal-manifold sampling, the ring angular velocity $\omega_z$ is drawn from $[\omega_{\min}, \omega_{\max}] = [1.5\pi, 3\pi]$\,rad/s.
We set $N_{\mathrm{planning}}=80$, $N_{\mathrm{smoothing}}=100$, $\Delta t_{\mathrm{col}}=1/30$\,s, and $f_{\mathrm{ctrl}}=240$\,Hz. 
For training the flow-matching model, we collect 60 executed trajectories by uniformly setting target distances from $[r_{\mathrm{exec}}^{\min}, r_{\mathrm{exec}}^{\max}] = [1.5, 2.0]$\,m in both simulation and real-world settings. 
Although the dataset is small, we will demonstrate in the experiments that the model effectively captures dynamics information and exhibits generalization capability. 
The difference between $[r_{\mathrm{plan}}^{\min}, r_{\mathrm{plan}}^{\max}]$ and $[r_{\mathrm{exec}}^{\min}, r_{\mathrm{exec}}^{\max}]$ reflects both safety considerations in real execution and the need to reserve part of the range for evaluating generalization. 

\subsubsection{Policy Learning Details}
We use $K=30$ basis functions, which provides a good balance between representation accuracy and model complexity.
The autoencoder maps each trajectory parameterization $\mathbf{p}_\tau$ into a $d_z$-dimensional latent space, with $d_z=64$ in our implementation. 
Both the encoder and decoder are 3-layer MLPs with hidden dimensions [256, 512, 256], and Leaky ReLU activations. 
The model is trained for up to 30,000 epochs with early stopping, using batch size 256 and Adam optimizer (learning rate $1\times 10^{-4}$, weight decay $1\times 10^{-5}$). 
The learned latent space then serves as input for the conditional flow-matching model. 
For this model, the network is a 6-layer MLP with hidden dimensions [256, 512, 1024, 1024, 512, 256] and Swish activations. 
Training uses batch size 450 with Adam optimizer (learning rate $3\times 10^{-4}$, weight decay $1\times 10^{-6}$) for up to 20,000 epochs. 
During trajectory generation, we integrate the learned vector field with the midpoint method using a step size of 0.001. 
All neural network inputs are normalized by subtracting the mean and dividing by the standard deviation. 



\begin{figure}[t]
    \centering
    \includegraphics[width=\linewidth]{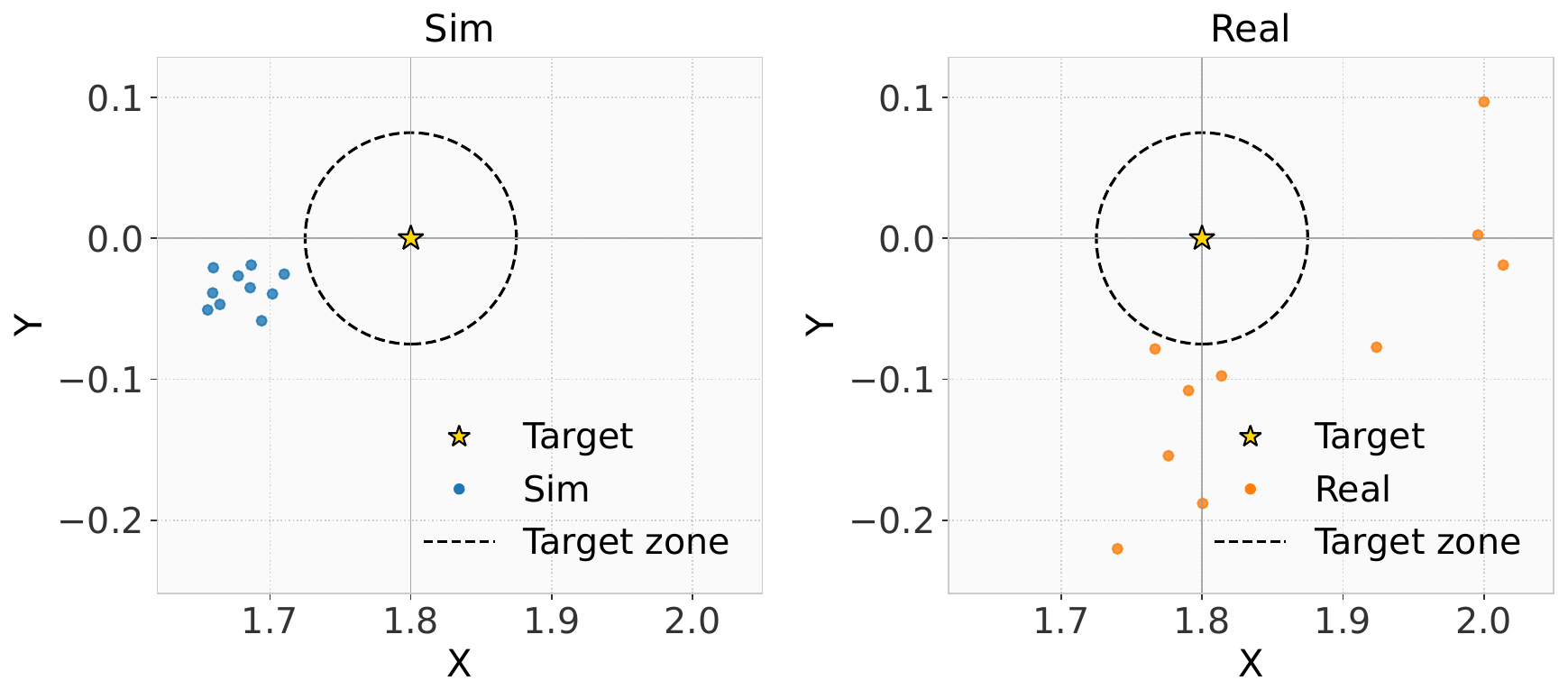}
    \caption{
    Comparison of landing distributions between simulation and the real world using motion planning (Baseline).
    Ten trajectories are planned targeting $r=1.8$\,m, and the actual landing positions are plotted.
    Note that the significant deviation and variance in the real world (orange dots) illustrate the dynamics gap that our proposed DA-MMP aims to close.
    This corresponds to the low success rate of the baseline reported in Table~\ref{tab:dynamics_aware}.
    }
    \label{fig:landing_scatter}
    \vspace{-10pt}
\end{figure}

\subsection{Dynamics-Aware Trajectory Generation}
To evaluate the effectiveness of our method in generating accurate throwing trajectories under dynamics gap, we compare it against multiple baselines in both simulation and real-world experiments, where in simulation we additionally introduce air drag by setting linear and angular damping.
We consider three baselines in simulation: motion planning with a single attempt, motion planning with two attempts, and a residual-style correction method that replans the second throw by compensating for the error observed in the first attempt, which is applied only when the first throw is unsuccessful.
In the real world, we further include two human baselines: novice participants throwing without any prior practice, and expert participants who have practiced for 60 trials before evaluation.
We report success rate (SR) as the evaluation metric, defined as the fraction of trials in which the ring successfully lands around the target peg. All results are averaged over three random seeds, each tested on 10 randomly sampled target positions.

The success rates of all methods are summarized in Table~\ref{tab:dynamics_aware}. 
Residual-style correction performs best in simulation but fails in the real world. 
Our method achieves the highest success rate in reality. 
Interestingly, it slightly outperforms trained human experts given the same number of practice trials, highlighting the efficiency of our framework which learns dynamics on top of structured motion primitives.

\begin{figure}[t]
    \centering
    \includegraphics[width=\linewidth]{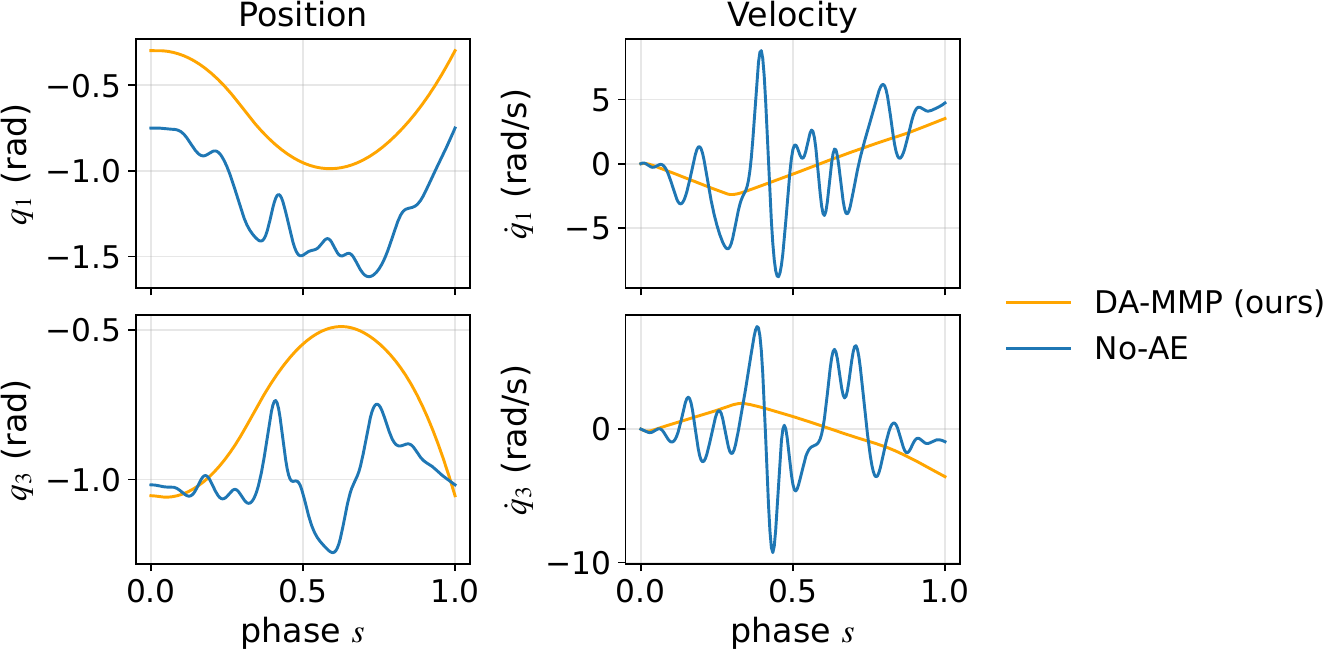}
    \caption{
    \textbf{Effect of the autoencoder on trajectory generation.}
    Shown are positions ($q_1, q_3$) and velocities ($\dot q_1, \dot q_3$) of two representative joints.
    Our method (orange) produces smoother and more coherent trajectories, whereas the no-AE variant (blue) exhibits irregular and oscillatory motions.
    }
    \label{fig:ablation_ae}
\end{figure}

To analyze the result reported in Table~\ref{tab:dynamics_aware}, we investigate three key questions:
\begin{enumerate}
\item Why does residual-style correction excel in simulation but fail in the real world?
\item Why does our method achieve a high success rate in both the simulation and the real world?
\item Can our method generalize to target distances not covered in the training data?
\end{enumerate}

To answer these questions, we visualize the landing distributions of motion planned throws in both simulation and the real world, as shown in Fig.~\ref{fig:landing_scatter}. 
In this evaluation, we plan ten trajectories targeting $r=1.8$\,m and plot the actual landing positions of the executed throws.
In simulation, the landing points exhibit a consistent bias with minimal variance, as the discrepancy is dominated by the added air drag rather than control or contact uncertainties.
In contrast, the real-world distribution exhibits both bias and variance due to additional control and contact dynamics gaps as well as visual perception errors.
This explains why residual-style correction is highly effective in simulation, as a systematic bias can be compensated for by adjusting the target. However, it fails in the real world because stochastic variance cannot be corrected in the same way (Question 1).
In fact, its performance (6.7\%) is even less than the marginal improvement gained by simply replanning a second time (23.3\% - 13.3\% = 10.0\%), as the compensation may even exacerbate the error when the variance is high. 
Our method, however, encodes trajectories rather than target positions, enabling it to capture trajectory-level dynamics and remain effective under real-world variability (Question 2).
Finally, as for Question 3, our method demonstrates promising generalization capability. 
When attempting throws to a distance of 1.2\,m, for which no real-world trajectories are collected—the policy generates feasible motions that successfully hit the target (as shown in the supplementary video). 
This indicates that the model learns the underlying trajectory–dynamics mapping, rather than merely memorizing training samples.

\begin{table}[t]
\centering
\caption{Effect of dataset scale on autoencoder reconstruction.
RMSE is measured in the normalized trajectory parameter space, and LRE is defined in (\ref{eq:lre}).}
\renewcommand{\arraystretch}{1.1}
\setlength{\tabcolsep}{8pt}
\begin{tabular}{ccc}
\Xhline{1.2pt}
\textbf{Dataset size} & \textbf{Parameter-Space RMSE} & \textbf{LRE (\%)} \\
\hline
0.09k  & 0.201 & 12.4 \\
0.9k  & 0.007 & 1.9 \\
9k    & 0.007 & 1.1  \\
90k   & \textbf{0.001} & \textbf{0.9}  \\
\Xhline{1.2pt}
\end{tabular}
\label{tab:ablation_scale}
\vspace{-10pt}
\end{table}

\subsection{Ablations}
\subsubsection{Effect of the Autoencoder}
To evaluate the role of the autoencoder in facilitating efficient training of the downstream generative model, we compare our approach against training the flow-matching model directly on raw trajectory parameterizations.
This variant lacks the manifold regularization provided by the embedding stage and therefore must model high-dimensional inputs directly.
On this test, we generate trajectories in an unconditional manner.
As shown in Fig.~\ref{fig:ablation_ae}, the non-embedded variant yields irregular joint motion profiles that are rarely executable, whereas our method generates smooth and feasible trajectories.
This demonstrates that the autoencoder is essential for capturing the intrinsic structure of feasible trajectories
and for mapping trajectories with their underlying dynamics.

\subsubsection{Effect of Dataset Scale}
We further study the effect of dataset scale on the quality of the learned motion manifold.
The autoencoder is trained with four different dataset sizes: 0.09k, 0.9k, 9k, and the full 90k planned trajectories,
with the network capacity reduced proportionally for smaller datasets to mitigate overfitting.
For evaluation, we sample 100 trajectories with a different random seed as a held-out test set.
We report two metrics: the normalized parameter-space reconstruction error (RMSE), and the relative length reconstruction error
\begin{equation}
    \mathrm{LRE} = \frac{1}{N}\sum_{j=1}^{N} \frac{|L^{(j)} - \hat L^{(j)}|}{L^{(j)}} \times 100\% .
    \label{eq:lre}
\end{equation}
where $L^{(j)}$ and $\hat L^{(j)}$ denote the ground-truth and the reconstructed trajectory lengths.
As summarized in Table~\ref{tab:ablation_scale}, larger training sets consistently produce lower RMSE and LRE, demonstrating that a sufficiently large corpus of planned trajectories is necessary to capture the diversity of feasible motions and establish a high-quality manifold that generalizes across targets.

\subsubsection{Effect of Radial Basis Functions}
To evaluate the utility of radial basis functions, we compare our parameterization against a linear interpolation baseline using $N_{\mathrm{wp}} = 32$ uniform waypoints and length $L$, matching our method's parameter count. We train autoencoders for both representations on three dataset scales (0.9k, 9k and 90k) and evaluate the reconstructed geometric smoothness on 100 held-out trajectories.

To assess geometric smoothness independent of execution duration, we sample $N_{\mathrm{eval}} = 100$ uniform points along the normalized phase $s \in [0,1]$ and approximate the second derivative at each interior point $s_i$:
\begin{equation}
    \mathbf{q}''(s_i) = \frac{\mathbf{q}(s_{i+1}) - 2\mathbf{q}(s_i) + \mathbf{q}(s_{i-1})}{\Delta s^2}
\end{equation}
where $s_i = \frac{i-1}{N_{\mathrm{eval}}-1}$ and $\Delta s = \frac{1}{N_{\mathrm{eval}}-1}$. The Mean Squared Second Derivative (MSSD) is then computed as:
\begin{equation}
    \text{MSSD} = \frac{1}{D(N_{\mathrm{eval}}-2)} \sum_{i=2}^{N_{\mathrm{eval}}-1} \left\| \mathbf{q}''(s_i) \right\|_2^2
\end{equation}
where $D=6$ is the robot's degrees of freedom. A lower MSSD indicates a smoother curve.

As shown in Table~\ref{tab:ablation_rbf}, two key conclusions emerge. First, our formulation yields significantly lower MSSD across all dataset scales, demonstrating that its inherent $C^2$ continuity acts as a strong structural prior to guarantee geometric smoothness. Second, for the waypoint-based baseline, trajectory smoothness noticeably improves as the dataset size increases (MSSD drops from 596.4 to 555.9). This indicates that training on a larger dataset helps the autoencoder learn a better latent representation, thereby mitigating high-frequency spatial reconstruction noise. Nevertheless, even with the largest dataset, the waypoint approach falls short of the smoothness naturally enforced by our radial basis parameterization, making it inherently suited for high-speed dynamic tasks.

\begin{table}[t]
\centering
\caption{
Effect of parameterization and dataset scale on geometric smoothness over 100 held-out trajectories.
}
\renewcommand{\arraystretch}{1.1}
\setlength{\tabcolsep}{8pt}
\begin{tabular}{lccc}
\Xhline{1.2pt}
\textbf{Parameterization} & \multicolumn{3}{c}{\textbf{MSSD vs. Dataset Size}} \\
\cline{2-4}
& 0.9k & 9k & 90k \\
\hline
Waypoints & 596.4 & 558.9 & 555.9 \\
DA-MMP (Ours) & \textbf{280.7} & \textbf{282.4} & \textbf{296.3} \\
\Xhline{1.2pt}
\end{tabular}
\label{tab:ablation_rbf}
\vspace{-10pt}
\end{table}

\section{CONCLUSION}

In this work, we present DA-MMP, a framework that builds upon motion manifold primitives for complex goal-conditioned dynamic manipulation tasks.
Our approach extends motion manifold primitives to variable-length trajectories through a compact parameterization, and leverages a large-scale dataset of planned trajectories to learn a high-quality manifold. 
On top of this manifold, we introduce a conditional flow matching model in the latent space that captures the relationship between trajectories and their execution dynamics, while relying on only limited real-world trials.
Our experimental evaluation confirms that DA-MMP consistently produces smooth and coordinated throwing motions with high real-world success rates.
Future directions include extending the framework to generalize across objects with diverse shapes and physical properties, exploiting its expressive motion generation to control object orientation and spin at release, and applying it to a broader range of goal-conditioned dynamic manipulation tasks.



\section*{ACKNOWLEDGMENT}

The authors would like to thank Zhecheng Yuan, Kun Lei, Qianwei Han, Xueyi Liu, and Lingxiao Guo for hardware assistance, Kaizhe Hu, Yu Qi for idea discussions, and Prof. Kris Hauser and Dr. Tobias Kunz for helpful discussions on kinodynamic planning algorithms. This work is supported by Tsinghua University Dushi Program.


\bibliographystyle{IEEEtran}
\bibliography{refs}

\end{document}